# ASR in German: A Detailed Error Analysis

Johannes Wirth
Research Group System Integration
Hof University of Applied Sciences
Hof, Germany
Johannes.Wirth.3@iisys.de

René Peinl
Research Group System Integration
Hof University of Applied Sciences
Hof, Germany
Rene.Peinl@iisys.de

*Abstract*— The amount of freely available systems for automatic speech recognition (ASR) based on neural networks is growing steadily, with equally increasingly reliable predictions. However, the evaluation of trained models is typically exclusively based on statistical metrics such as WER or CER, which do not provide any insight into the nature or impact of the errors produced when predicting transcripts from speech input. This work presents a selection of ASR model architectures that are pretrained on the German language and evaluates them on a benchmark of diverse test datasets. It identifies cross-architectural prediction errors, classifies those into categories and traces the sources of errors per category back into training data as well as other sources. Finally, it discusses solutions in order to create qualitatively better training datasets and more robust ASR systems.

## I. Introduction

"Benchmarks that are nearing or have reached saturation are problematic, since either they cannot be used for measuring and steering progress any longer, or – perhaps even more problematic – they see continued use but become misleading measures: actual progress of model capabilities is not properly reflected, statistical significance of differences in model performance is more difficult to achieve, and remaining progress becomes increasingly driven by over-optimization for specific benchmark characteristics that are not generalizable to other data distributions"[1]. Hence, novel benchmarks need to be created to complement or replace older benchmarks (ibid.). This description is also true for the English automatic speech recognition (ASR) benchmark Librispeech as seen in Figure 1. Therefore, several different benchmarks have been proposed, including Switchboard, TIMIT and WSJ. However, even on those, a couple of systems already achieve low single digit word error rates (WER) and small improvements are hard to interpret [2]. In the case of Switchboard, a considerable portion of the remaining errors involves filler words, hesitations and non-verbal backchannel cues (ibid.).

This paper investigates whether benchmarks for German ASR exhibit similar saturation, although there is far less available data for the German language compared to English with correspondingly fewer benchmark results being published. This includes detailed error analysis and error backtracing to determine specific improvements that are necessary to increase model performance in terms of training data quality and to more realistically represent model performance in evaluations.

The rest of the paper is structured as follows. Related work is presented in section 2. Section 3 presents an overview of ASR models considered and which ones were chosen for further analysis. Section 4 describes the datasets that were used for analysis. Results in terms of WER are shown in section 5 before the detailed error analysis is presented in section 6. We briefly discuss possible resolutions for the different kinds of error in section 7 before the conclusion and outlook in section 8.

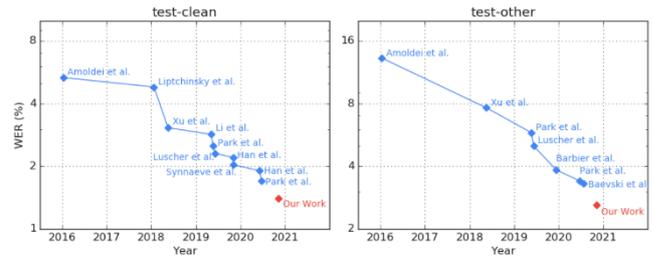

Figure 1: Librispeech benchmark reaching saturation [3]

## II. Related Work

Previous work was done on analyzing the robustness of ASR models in English language, by running them through a diverse set of test data [4], [5], see Table 1. Likhomanenko et al. found that for models trained on a specific dataset there is a huge discrepancy between the WER when evaluating them on a test split of the same dataset and evaluating them on other datasets. In other words, ASR models are often not able to generalize well and the authors demand that models should always be evaluated on a diverse dataset consisting of at least Switchboard, Common Voice and TED-LIUM v3 for English language in the future. To reach that conclusion, they evaluated a single ASR model on five different public datasets.

TABLE I. BENCHMARKS USED FOR ASR EVALUATIONS

| Source | Datasets for evaluation |
|---|---|
| [4] EN | WSJ, LibriSpeech, Tedium, Switchboard, MCV |
| [5] EN | ST, LibriSpeech, Tedium, Timit, Voxforge, RT, AMI, MCV |
| [5] DE | Voxforge, MCV, Tuda, SC 10, Verbmobil II, Hempel |
| ours DE | Voxforge, MCV, Tuda, VoxPopuli, MLS, M-AILABS, SWC, HUI, German TED Talks, ALC, SL100, Thorsten, Bundestag |

Ulasik et al. collected a similar benchmark corpus of publicly available data for both English and German language, called CEASR. They evaluated it on four commercial and three open-source ASR systems. They found that the commercial systems outperform the open-source systems by a large percentage for some datasets. Additionally, the performance differences of single systems across different datasets were large. However, they use open-source systems that seem a bit outdated, even for the year of publication [6] and only conducted evaluations for the English language. For German, only anonymous commercial systems were evaluated. The best performing model achieves 11.8% WER on Mozilla Common Voice (MCV), which is less than Scribosermo, an open source model based on Quartznet that delivers 7.7% WER on MCV [7]. For Tuda, (see section IV) Scribosermo reports 11.7% WER whereas Ulasik et al. reports 13.05% for the best commercial system. In contrast to that, our work evaluates six state-of-the-art (SOTA) open-source models in German language on thirteen datasets.

Aksënova et al. analyzed the requirements for a new ASR benchmark, that fixes the shortcomings of existing ones [2]. They stated that a next-gen benchmark should cover the application areas dictation, voice search & control, audiobooks, voicemail, conversations and meetings as well as podcasts, movies and TV shows. Furthermore, it should cover technical challenges such as varying speed, acoustic environments, sample frequencies and terminology (ibid.). They identified coherent transcription conventions as a key element for building high quality datasets, as well as detecting and correcting errors in existing corpora. This paper applies exactly these approaches for the German language.

Keung et al. [8] show that modern ASR architectures may even start emitting repetitive, nonsensical transcriptions when faced with audio from a domain that was not covered at training time. We found similar issues in our analysis.

## III. ASR MODELS

There is a large variety of ASR models presented in literature [9]. Our selection considers speaker independent models with a very large vocabulary for continuous as well as spontaneous speech according to the classification in [9]. We are further looking for models with a publicly available implementation, either by the original authors of the model or a third party. Finally, we are using models that were already pretrained on German language. After first tests, we excluded models, that performed bad in our first tests without a language model, including Mozilla DeepSpeech [10] and Scribosermo [7], which are both available as part of the Coqui project[1].

The Kaldi project, which is one of the most popular open-source ASR projects, does not provide any publicly available pretrained models for German language[2]. The reported WERs for their model on MCV, Tuda, SWC and Voxforge are however worse than that of all models tested in this work, except Quartznet, although Tuda and SWC were included in the training data of Kaldi. Speechbrain [11] can be seen as a potential successor of Kaldi with a more modern architecture and focus on neural models. It provides a CRDNN with CTC/Attention trained on MCV 7.0 German on huggingface[3] with a self-reported WER of 15.37% on MCV7. CRDNN stands for an architecture that combines convolutional, recurrent and fully connected layers.

Most of the tested models are provided by Nvidia as part of the NeMo toolkit [12]. Quartznet and Citrinet share a very similar architecture. While Quartznet targets small lightweight models with fast inferencing [13], Citrinet is more targeted towards scaling to higher accuracy while keeping a moderate fast inferencing speed [14]. ContextNet was originally developed by Google [15] and was adopted by Nvidia to be included in their NeMo toolbox. Finally, we have evaluated two Conformer models [16] that have been trained by Nvidia, one that is using CTC loss and the other one being a Conformer Transducer [17]. All Nvidia models can be found online [4]. We always use the "large" models if different versions are available. All models from Nvidia except Quartznet are trained on the German parts of Mozilla Common Voice v7.0, Mulilingual LibriSpeech and VoxPopuli. Quartznet has been pretrained on 3,000 hours of English data and then finetuned on Mozilla Common Voice v6.0 (see TABLE II. ).

TABLE II. ASR MODELS WITH TRAINING DATA EVALUATED

| Model | Datasets for training |
|---|---|
| Citrinet | German MCV 7, MLS, VoxPopuli |
| ContextNet | German MCV 7, MLS, VoxPopuli |
| Conformer CTC | German MCV 7, MLS, VoxPopuli |
| Conformer Transducer | German MCV 7, MLS, VoxPopuli |
| Wav2Vec 2.0 | Pretraining on MLS EN, finetuning on MCV6 |
| Quartznet | Pretraining on 3000h EN, finetuning MCV6 |

The last model evaluated in detail is Wav2Vec 2.0 [3] from Facebook AI research (FAIR). There are several pretrained models available on Huggingface. We picked the Wav2Vec2-Large-XLSR-53-German by Jonatas Grosman [18], since it had the lowest self-reported WER on Common Voice (12.06%) compared to 18.5% reported for the original model provided by FAIR. Recently, there was an update on the model [19] with the new MCV v8 data, which was not included in the evaluation.

All models were evaluated without utilizing any language model in order not to introduce another source of error into the analysis. In this context, it is worth noting that two of the six models evaluated, ContextNet and Conformer Transducer, are autoregressive (based on RNN-T loss) while all others are based on CTC loss.

## IV. DATASETS

The evaluation was intended to include as many and diverse publicly available datasets as possible. They should include spontaneous as well as continuous speech, shorter and longer sentences as well as a diverse set of vocabulary in order to find systematic errors that models have learned and to identify how well the selected models can generalize on unseen data.

Mozilla Common Voice [20] is a multi-lingual speech corpus with regular releases and a steadily growing amount of data. It is crowd sourced and contains 1,062 hours of German speech data in its 8.0 release from January 2022 (965 h for v7.0). Tuda is a dataset for German ASR that was collected by TU Darmstadt in 2015 [21]. In contrast to those, some of the datasets collected from the Bavarian Speech Archive (BAS) were not intended for ASR but "were originally intended either for research into the (dialectal) variation of German or for studies in conversation analysis and related fields" [22]. Tuda contains 127 hours of training data from 147 speakers with very little noise.

Several datasets are based on the Librivox project [23], where volunteers read and record public domain books. Among them is the M-AILabs dataset [24] which contains 237 hours of German speech from 29 speakers. It also includes Angela Merkel, a speaker from a political context and thus quite different from the other speakers. We therefore excluded this part from the M-AILabs dataset and listed it separately. Voxforge is another dataset derived from Librivox [10]. It contains only 35 hours of speech from 180 speakers. MLS is the Multilingual LibriSpeech corpus [25] compiled by Facebook that contains over 3,000 hours of German speech and is also based on Librivox. Finally, the HUI speech corpus was originally designed to be used for text-to-speech (TTS) systems [26], but can also be used for ASR with its 326 hours of German speech recorded by 122 different speakers of the Librivox project. We expect the datasets based on Librivox to have a large overlap; however, no further analyses were conducted to determine exact values.

---

[1] https://coqui.ai/models
[2] https://github.com/german-asr/kaldi-german
[3] https://huggingface.co/speechbrain/asr-crdnn-commonvoice-de
[4] https://catalog.ngc.nvidia.com/orgs/nvidia/collections/nemo_asr

Similar to the HUI dataset is Thorsten Voice [27]. It also was created for training TTS systems with 23 hours of German speech. It was, however, recorded by a single speaker. It contains mostly short sentences, primarily related to voice assistant prompts. All spoken words and utterances are pronounced in an unambiguous and very clear manner.

Another major source of audio data in German originates from political speeches. VoxPopuli is a multilingual dataset based on speeches held at the European parliament [28]. It contains both unlabeled and labeled data, which results in 268 hours of German speech data suitable for training and evaluation. At Hof University, an ASR dataset based on speeches at the German Bundestag[5] was prepared using 211 hours of commission sessions and 393 hours of plenary sessions. The test split of the dataset is already released[6].

The Spoken Wikipedia Corpus (SWC) is a multilingual dataset comprising 285 hours of read articles from Wikipedia [29]. "Being recorded by volunteers reading complete articles, the data represents the way a user naturally speaks very well, arguably better than a controlled recording in a lab. The vocabulary is quite large due to the encyclopedic nature of the articles" [30].

One of the few datasets with spontaneous speech besides the political speeches is the TED Talks corpus [31]. It is similar to the English TED-lium corpus [32] and contains data from German, Swiss and Austrian speakers. It contains only 16 hours of speech but is a valuable addition due to its unique features.

Additionally, we also included a few corpora that were collected for linguistic research, namely ALC, a corpus for the use of language under the influence of alcohol [33] and BAS SI100, a corpus created by the LMU Munich with 101 speakers (50 female, 50 male, 1 unknown). Each speaker has read ~100 sentences from either the SZ subcorpus or the CeBit subcorpus [34]. The subcorpus SZ contains 544 sentences from newspaper articles ("Sueddeutsche Zeitung"). The subcorpus CeBit contains 483 sentences from newspaper articles about the CeBit 1995.

TABLE III. summarizes the key metrics of the datasets. The numbers reported for M-AILabs are for the whole dataset regarding number of speakers and total hours, but for test hours and durations we are referring to the test dataset without Merkel. The statistics of this subset are reported separately. We used the official test splits where applicable. For the other datasets, a random split with 10% of the overall data per set was created. Inclusion of more datasets proved to be unnecessarily tedious due to diverse formats of transcriptions with no tool to automatically convert between different styles.

## V. EVALUATION RESULTS

First, a conventional evaluation of the models based on WER was performed using the WER collected on every dataset in order to put their accuracy into perspective and to collect prediction errors for further processing.

TABLE III. DATASETS USED IN EVALUATION

| dataset | hours (all \| test) | speakers | type | len (min/ø/max) |
|---|---|---|---|---|
| MCV 7.0 | 965 \| 26.8 | 15,620 | read | 1.3/6.1/11.2s |
| Tuda | 127 \| 11.9 | 147 | read | 2.5/8.4/33.1s |
| SWC | 285 \| 9.1 | 363 | read | 5.0/7.9/24.9s |
| M-AILabs | 237 \| 10.8 | 29 | read | 0.4/7.2/24.2s |
| MLS | 3287 \| 14.3 | 244 | read | 10/15.2/22s |
| VoxForge | 35 \| 2.7 | 180 | read | 1.2/5.1/17.0s |
| HUI | 326 \| 16.3 | 122 | read | 5.0/9.0/34.3s |
| Thorsten | 23 \| 1.1 | 1 | read | 0.2/3.4/11.5s |
| VoxPopuli | 268 \| 4.9 | 530 | spoken | 0.6/9.0/36.4s |
| Bundestag | 604 \| 5.1 | | spoken | 5.0/7.2/35.6s |
| Merkel | 1.0 | 1 | spoken | 0.7/6.9/17.1s |
| TED Talks | 16 \| 1.6 | 71 | spoken | 0.2/5.1/118s |
| ALC | 95 \| 2.6 | | read | 2.0/12.5/62s |
| BAS SI100 | 31 \| 1.8 | 101 | read | 2.1/12.7/54.1s |

TABLE IV. WORD ERROR RATES FOR ALL MODELS AND ALL DATASETS

| | Citrinet | Conf. CTC | Conf. T | Contextnet | Wav2Vec 2.0 | Quartznet |
|---|---|---|---|---|---|---|
| **MCV 7.0** | 8.78% | 8.00% | **6.28%** | 7.33% | 10.97% | 13.90% |
| **Bundestag** | 13.25% | 13.65% | **11.16%** | 14.44% | 21.78% | 28.61% |
| **VoxPopuli** | 10.35% | 10.82% | **8.98%** | 10.13% | 21.96% | 28.34% |
| **Merkel** | 13.63% | 17.17% | **13.49%** | 15.92% | 21.81% | 27.57% |
| **MLS** | 5.56% | 5.16% | **4.11%** | 4.62% | 13.04% | 20.34% |
| **MAI-LABS** | 5.52% | 5.56% | **4.28%** | 4.32% | 9.94% | 18.47% |
| **Voxforge** | 4.15% | 3.95% | **3.36%** | 4.16% | 5.64% | 7.58% |
| **HUI** | 2.31% | 2.45% | **1.89%** | 2.02% | 8.52% | 14.66% |
| **Thorsten** | 6.74% | 8.49% | 6.20% | 9.21% | 7.57% | **5.95%** |
| **Tuda** | 9.16% | 7.81% | **5.82%** | 7.91% | 12.69% | 20.31% |
| **SWC** | 10.15% | 9.36% | **8.04%** | 9.29% | 15.01% | 16.49% |
| **German TED** | 34.53% | 35.77% | **31.98%** | 35.58% | 41.90% | 47.75% |
| **ALC** | 31.42% | 31.30% | **25.90%** | 26.85% | 40.94% | 45.53% |
| **BAS SL100** | 23.13% | 24.81% | 22.82% | 22.74% | 28.84% | 28.94% |
| **Average** | 12.76% | 13.16% | **11.02%** | 12.47% | 18.62% | 23.17% |
| **Median** | 9.65% | 8.93% | **7.16%** | 9.25% | 14.02% | 20.33% |

The numbers reported in TABLE IV. show that Conformer Transducer outperforms all other models regarding WER. Quartznet scored worst on all but a single test set. Citrinet, Conformer CTC and ContextNet perform very similar with less than 0.5% absolute difference in WER. However, the median of Conformer CTC is the best of those three models although the average is the worst, which means that its output predictions show higher variation than those of the two other models, which we count as an indicator for lower robustness. Citrinet and Quartznet have the lowest difference between average and median, which we count as an indicator for robustness. Wav2Vec 2.0 is the second worst model and does exceptionally bad for German TED and ALC. This is similar to Quartznet. We hypothesize that this stems from filling words like "äh" and "hm" occurring in their output transcript predictions, which, in contrast, are omitted in the predictions of all other models. These fillers are also missing in all ground truth transcripts. The WERs for the three political datasets are nearly identical for Wav2Vec. For the Conformer models and Contextnet, VoxPopuli WERs are much better than for Bundestag and Merkel, which is expected with respect to the training data. However, it is not clear why Bundestag has lower WERs than Merkel for those models. It could be due to less errors in the transcript. The HUI dataset has the lowest WERs of all datasets. This is due to the alignment process, where ASR was already used and sentences, that could not be aligned with ASR were discarded. The significant differences in WERs between Merkel and the rest of MAI-Labs on the one hand and the similarities to Bundestag on the other hand, confirm our decision to treat Merkel separately.

---

[5] https://www.bundestag.de/mediathek, last access: 28.03.2022

[6] https://opendata.iisys.de/

In general, the datasets with spontaneous speech show a higher WER than the ones with continuous speech, which is in line with findings for the English language [4].

TABLE V. PERFORMANCE COMPARISON OF ASR MODELS

|  | # of params | size on disk | autoregressive | RTF | speed |
|---|---|---|---|---|---|
| Citrinet | 142.0 Mio | 532 MB | no | 0.007 | 158% |
| Conformer CTC | 118.8 Mio | 452 MB | no | 0.006 | 124% |
| Conformer T | 118.0 Mio | 446 MB | yes | 0.015 | 323% |
| Contextnet | 112.7 Mio | 476 MB | yes | 0.013 | 291% |
| Wav2Vec 2.0 | 317.0 Mio | 1204 MB | no | 0.021 | 464% |
| Quartznet | 18.9 Mio | 71 MB | no | 0.005 | 100% |

When evaluating the speed of the models in a batch operation with ~1h of audio in 600 files on a AMD Ryzen 3700X with 16 GB RAM and Nvidia GTX 1080 GPU, they all perform reasonably well (see **Fehler! Verweisquelle konnte nicht gefunden werden.**). Quartznet is the fastest, with Conformer CTC and Citrinet being close. ContextNet and Conformer Transducer are about three times slower due to their autoregressive nature. Wav2Vec 2.0 performs worst due to its large model size and needs 4.6 times as long as Quartznet.

## VI. ERROR IDENTIFICATION

As shown in TABLE IV. , WER can be used to determine how well models perform generally and in relation to each other. However, a single metric does not suffice for the identification of cases, in which ASR systems perform particularly poorly and for what reasons. The following sections describe the methods performed to accumulate cross-model and model-exclusive prediction errors, to classify those further, and how their root causes can be traced back to training and test data.

### A. Method

For deeper error analysis, subsets of the total error set (all predictions with WER ≠ 0, excluding pure spacing errors) of all models were created. On one hand, difference sets were formed to identify model- and architecture-specific error sources. On the other hand, the intersection of the incorrectly predicted transcripts of all models was created to identify model-independent error causes, which this work focuses on. It was expected that consistent cross-model errors could be traced back to shared training data or ambiguous and low-quality test data.

In addition, the vocabulary of the training datasets of the models was accumulated and compared to that of the test datasets in order to determine how well models can generalize to unseen words, as well as how many words were regularly predicted incorrectly although they were contained within the training datasets.

### B. Error Classification

From the intersection of the previously described cross-model prediction errors, 2,000 samples were extracted and manually assigned error categories. This partitioning was intended to more distinctively specify the degree of impact and the cause of several groups of errors in false transcript predictions. All mispredicted transcripts across all models could be assigned at least one of the following categories.

*1) Negligible Errors*

This category includes different forms of otherwise correct transcript predictions, i.e., most models in use were trained on nonnormalized and commonly abbreviated phrases like "et cetera" or "etc.", which increases the word error rate due to normalized ground truth transcripts but does not represent an actual error. All models except Quartznet and Wav2Vec 2.0 were trained using nonnormalized abbreviations, which appear in the training datasets of MLS and VoxPopuli. This category also includes words which can be correctly written in multiple possible ways (including obsolete orthography).

*2) Minor Errors (noncontext-breaking)*

Primarily Quartznet and Wav2Vec 2.0, which were trained on less German data than all other models, often produce transcript predictions with redundant letters, omit single letters or predict hard instead of soft vowels and vice versa (e.g., confusion between d and t) without distorting the meaning. These errors are quickly recognized by humans as spelling errors and can usually be corrected when utilizing a language model, thus having little impact on potential context misinterpretations of model outputs. While increasing character error rate only slightly, WER quickly rises to unrealistic values if these minor errors are included.

*3) Major Errors (context-breaking)*

Fully incorrectly transcribed or omitted words and omitted or inserted letters that change the meaning of a transcript or exclude necessary information and sentence components represent the most critical category for which the causes were to be identified. Those are further discussed in the following section. A specifically humorous example is:

- Ground truth: "an einem st**and** werden waff**eln** verkauft um die vereinskasse aufzubessern" (EN: waffles sold)

- Prediction: "an einem st**ra**nd werden waff**en** verkauft um die vereinskasse aufzubessern" (EN: weapons sold)

*4) Names, Loan Words and Anglicisms*

A commonly occurring type of error in ASR systems, since names can often be spelled in several ways and only a small fraction of common names is usually found in training datasets, foreign words are strongly domain dependent, and some anglicisms have homophonic pronunciations to German phonemes and therefore can often only be correctly interpreted in a strongly context-dependent manner. Anglicisms are closely related to code-switching ASR.

Although errors in this subcategory often have similar effects to contextual understanding as major errors, they have been separated due to being generally underrepresented in publicly available datasets for a single language.

*5) Homophones*

Another highly contextual source of error, which is often obviously recognized as wrong by humans, but can change the entire context of transcripts in some cases (e.g., "Graph" and "Graf"; in English: "Graph" and "Count", a noble title). In those instances, contextual information is lost, but the underlying phonemes were in general correctly interpreted by the system. Occurrences of homophones might also indicate that a network was merely trained on a single representation and context of phonemes, thus always predicting the previously seen representation.

*6) Flawed Ground Truth Transcripts*

While it is assumed that almost all transcripts correspond to the actual statements of the respective audio recordings in training and test data sets, there are frequent inconsistencies. This is caused by various factors such as incorrect normalization of numbers and symbols, transcripts that are correct in terms of their meaning but not in terms of their wording, and strong deviations and tolerances in the alignment process during the (automated) creation of data sets.

TABLE VI.    EXEMPLARY COMPARISON BETWEEN GROUND TRUTH, ACTUAL TRANSCRIPT AND TRANSCRIPT PREDICTIONS

| Ground truth | auf der bruecke **warteten** der koenig und der kronprinz |
|---|---|
| Manual transcript | auf der bruecke **warten** der koenig und der kronprinz |
| All models | auf der bruecke **warten** der koenig und der kronprinz |

TABLE VI. shows a typical example of erroneous ground truth transcripts. While all ASR systems predicted the word "warten", which was verified by a manual transcription, the actual ground truth transcript contains the word "warteten". These incorrect data consequently lead to worse benchmark results of the models than deserved.

*7) Ambiguous Audio Input*

Especially in spontaneously spoken utterances, dialogues, and uncontrolled recording conditions, but also through stuttering, certain words may be pronounced or perceived indistinctly. This can result in transcript predictions of homophones, but also complete word substitutions or omitted words.

*8) Flawed Audio Input*

Especially in automatically generated datasets (e.g. by utilization of CTC segmentation [35]) long audio recordings are often split very precisely at the end of a word, which can lead to cutoffs of spoken words at the beginning or end of audio snippets. In addition, datasets such as MCV and Voxforge are not fully quality-assured and contain recordings with extremely poor recording quality, which are difficult to understand, even for human listeners. These aspects form a category of their own, as it cannot be expected that correct transcript predictions can be generated from incomplete or corrupted input data.

## C. Results and Error Origins

After manually transcribing and classifying 2,000 audio samples with incorrect transcript prediction across all used models, the numbers of occurring errors were put in relation as displayed in TABLE VII.

TABLE VII.    ERROR CLASSIFICATION AND PROPORTIONS

| Nr | Error Category | Error Proportion |
|---|---|---|
| 1 | Negligible | 9,40% |
| 2 | Noncontext-Breaking | 11,95% |
| 3 | Context-Breaking | 19,01% |
| 4 | Name, Anglicism, Loan Word | 19,82% |
| 5 | Homophone | 2,92% |
| 6 | Flawed Ground Truth Transcript | 17,85% |
| 7 | Ambiguous Audio Input | 11,13% |
| 8 | Flawed Audio Input | 7,91% |

Within the classified samples, about 27% of prediction errors (category 1+6) can be attributed to incorrect ground truth transcripts, outdated spellings and mostly negligible word separation errors, from which can be concluded, that the realistic WER across models would be even lower than previously determined by automated WER calculations. This proportion would rise even further if flawed audio inputs were also considered.

With just under one fifth of the classified errors, names, foreign words and anglicisms make up a large part of all cross-model mispredictions. This type of misprediction ranges from alternative spellings for names (with little effect) to completely incomprehensible transcript predictions, which have a similar effect on human comprehension as errors from the context-breaking category. However, this category is generally known to cause problems for ASR systems. Humans also have problems correctly spelling names, they are not familiar with. A common way to obtain more reliable predictions is finetuning on domain-specific datasets and utilizing a language model.

In more than 11% of the samples (category 7), even human listeners were unable to produce clear and correct transcripts from audio recordings, because the pronunciation was too unclear. This could be traced back to speakers having learned German as a second language (L2 speakers), multiple German dialects, simple pronunciation errors and slips of the tongue. Increasing the robustness of ASR systems for non-native speakers and dialects is an ongoing research topic (e.g., [36]) and is especially challenging due to lack of standard orthography [37] for dialect speech.

Nevertheless, almost another fifth of the total errors contained in the sample derive from words consistently predicted incorrectly or mapped to entirely different words, some with similar meaning, some completely incorrect. Because this category has the strongest impact on reading comprehension, these errors were further divided into subgroups and their causes traced back to systematic errors within the training data. All the following observed errors occur exclusively in predictions from models trained on MLS and VoxPopuli (both datasets were prepared by FAIR):

*1) Naive Normalization*

The most obvious instance of incorrect normalization can be seen in the transcription of pronounced years from the past millennium. In this case, the training data sets do not contain

years written out but interpreted as integers. Consider the following example:

- Ground truth: Neunzehnhundertdreiundsechzig (Nineteen Sixty-Three)
- Prediction: Eintausendneunhundertdreiundsechzig (One Thousand Nine Hundred Sixty-Three).

All models except for Quartznet and Wav2vec 2.0 produced this misprediction.

Consistently incorrect normalization of numbers in training data does not only affect years, also larger integer values like "siebzigtausend" (seventy thousand) are predicted as "siebzig null" (seventy zero).

In addition, pronounced punctuation marks such as "punkt" (full stop) or "komma" (comma) were not converted to verbatim expressions in training data but completely removed and thus ignored in predictions. Lack of normalization was particularly evident in the ALC and SI100 test datasets, which intentionally contain many numbers (including chains of single digits) and literally pronounced punctuation marks. Therefore, one important distinction is whether corpora adopt 'spoken-domain' transcriptions (orthographic), where numbers are spelled out in words (e.g. 'three thirty'), or 'written-domain' transcriptions, where they are rendered in the typical written form ('3:30', nonorthographic) [2]. Models trained on one type of dataset will generate errors when evaluated with the other type of dataset.

*2) Indirect Transcription*

A significant number of ground truth transcripts contain statements of the same meaning, but not the verbatim expression of the corresponding audio snippet. For instance, the word "weil" (because) was interpreted regularly as "denn" (since), which did not change the meaning in any of the observed cases. Alteration of context cannot be ruled out for further samples.

*3) Consistently Wrong Transcriptions*

Certain terms such as "paragraph" (article) were transcribed as "ziffer" (subparagraph) within the ground truth transcripts. As a result, models trained on this wrong data consistently predicted these spoken words (as well as a number of others) incorrectly.

*4) Constant Sentence Structures*

The VoxPopuli dataset occasionally contains the statement "Herr Präsident" (Mister President) at the beginning of a sentence, but in several cases, this was not actually said; it is a systematic transcript error. Trained models have adopted this phrase and regularly predict it at sentence beginnings. This was primarily observed in predictions of test datasets with similar recording environments to VoxPopuli (e.g., TED talks). Additionally, for poorly edited audio inputs (starting in the middle of a sentence or word), models predicted sentence beginnings that were not present in the audio, and in most cases turned out to be incorrect. This type of error fits into the category " hallucinations" as described in [2].

The previously listed categories reveal systematic errors in the datasets MLS and VoxPopuli primarily caused by flawed data preprocessing steps, which become particularly obvious in contrast to transcript predictions by models trained without these datasets.

*D. Further Sources of Error*

While the error classification and analysis identified systematic dataset errors, it was also examined how well models could perform predictions for words outside the vocabulary of their training data.

TABLE VIII. TRAINING VOCABULARY, ABSOLUTE NUMBER OF UNSEEN WORDS AND CORRECTLY PREDICTED UNSEEN WORDS IN PERCENT

| Model | Training Vocabulary | Unseen Words | Well Generalized |
|---|---|---|---|
| Citrinet | 467,740 | 15,836 | 47% |
| Conformer CTC | 467,740 | 15,836 | 50% |
| Conformer T | 467,740 | 15,836 | 53% |
| Contextnet | 467,740 | 15,836 | 50% |
| Wav2Vec 2.0 | 210,407 | 32,972 | 52% |
| Quartznet | 210,407 | 32,972 | 45% |

TABLE VIII. shows that about half of all words that were unseen during the training process could be transcribed correctly. It should be noted that Quartznet and Wav2Vec 2.0 were trained using a significantly smaller German vocabulary. Despite that, Wav2Vec 2.0 was able to correctly predict an above-average number of words.

TABLE IX. WER ON THORSTEN WITH AND WITHOUT SILENCE

|  | original | with silence | delta |
|---|---|---|---|
| Citrinet | 6.74% | 4.09% | -2.65% |
| Conformer CTC | 8.49% | 3.96% | -4.53% |
| Conformer Trans | 6.20% | 3.61% | -2.59% |
| Contextnet | 9.21% | **3.35%** | -5.86% |
| Wav2Vec 2.0 | 7.57% | 5.11% | -2.46% |
| Quartznet | **5.95%** | 5.00% | -0.95% |

A final error that affects multiple datasets is especially severe for Thorsten (see TABLE IX. ). All models perform significantly worse on audio files that have too few leading and trailing silence included. This was already recognized in previous work [38]. After finding many missing characters or words at the beginning and end of the transcripts, we added 0.3s silence at the beginning and end of every audio file for Thorsten. Rerunning the ASR inferences showed a significant decrease in WERs so that Thorsten became the best recognized dataset after HUI, although it was not included in the training data like Voxforge. ContextNet benefits most of this additional silence with a WER reduction of 5.9% absolute, whereas Quartznet seems to be affected the least by missing silence. It took the lead regarding WER for the original Thorsten data and had only a minor reduction in WER with additional silence.

## VII. PROPOSED RESOLUTIONS

For the cross-model errors identified and evaluated in the previous chapter, a selection of methods was determined to minimize or eliminate the impact of these errors in future trained models.

*A. Verification of (correct) Normalization*

Since all models trained using MLS and VoxPopuli produce inconsistent output for spoken years, single letters, symbols, and abbreviations due to incorrect normalization of the training data, further verification of the correctness of these datasets must be conducted. The primary requirements are cross-dataset consistency of normalization and avoidance of information loss (e.g., no general removal of punctuation marks).

Performing a mostly automated verification of training datasets and manually correcting or discarding poorly normalized data rows can significantly increase the degree of alignment between audio recordings and transcripts, not only preventing incorrectly trained alignments from occurring within predictions, but potentially improving overall model performance through more consistent overall training of audio-transcript-alignment.

### B. Extension of Vocabulary through Text To Speech

By using a text-to-speech system, models could be fine-tuned on domain-specific terms for which no audio recordings exist. Furthermore, training data could be generated which does not correspond to any sentence structure of written texts and more to that of spontaneously spoken utterances. This could increase the performance of models in dialogues and natural speech as well as prevent architectures from overfitting on frequent beginnings of sentences.

However, it must be ensured that the TTS system itself generates sufficiently good predictions for the given words. Especially with terms and names of several different languages, insufficiently trained neural networks for speech synthesis may lead to unusable audio outputs and even worsen the performance of models for speech recognition.

### C. Training on Phoneme Vocabulary

Although the use of phoneme-based ASR systems adds another non-trivial step, the mapping of phonemes to character sequences, further measures can be applied to determine contextually correct homophones or to separately interpret and process the speech input of L2 speakers or speakers with dialect. This can be implemented using other specialized systems on, e.g., dialect classification [39], thus detaching more specialized and advanced tasks from speech recognition systems.

### D. Audio Preprocessing

Particularly for dialogues, podium speeches and recordings with poor recording quality, it may be worthwhile to apply various methods of audio preprocessing, such as speaker separation, audio normalization and audio enhancement. The utilization of these methods of course depends on whether the desired system should consist of a single, more robust ASR model, or a composition of preprocessing steps and a downstream ASR model. Undoubtedly, such measures degrade the real time factor of an ASR system but could significantly increase the quality of transcript predictions or assist in the identification of flawed audio inputs.

## VIII. CONCLUSION

The presented work has demonstrated that sources of error of speech recognition systems based on neural networks can be identified by the generation of sets of prediction errors as well as by dataset and vocabulary analysis, and that those can be traced back to consistently erroneous training and test data. By utilizing the mentioned methods, it was not only shown that models perform on average better than represented by metrics such as automatically calculated WER on test benchmarks, but also how erroneous data preparation and other task-specific aspects have a lasting negative impact on the performance of trained models.

Implementation of the countermeasures proposed for this purpose could lead to more realistic evaluations in future models, as well as more robust models overall. However, some aspects of automatic speech recognition are still weak.

It turns out, that there is no single ASR model, that works best in all scenarios. Although Conformer Transducer has the lowest WER overall, it produces a number of errors that are prohibitive in some scenarios. Especially the described hallucinations, omitting and changing of words impact transcript predictions very negatively, if one is interested in the details a person said. Wav2Vec 2.0 on the other hand has a comparably high WER overall but stays very close to the phones that were really said.

Since there are no large scale ASR datasets in German language like SPGISpeech [40] with 5,000 hours or Gigaspeech [41] with even 10,000 hours of transcribed audio, future research should aim at carefully correcting errors in existing corpora and complement them with additional data that fills the gaps, like the underrepresented spontaneous speech. It would also be desirable to have several alternative versions of the transcripts so that users can choose the style that is suitable for their application area. There should further be a GDPR-compliant large-scale dataset for German ASR, at least with unsupervised data like LibriLight. Additionally, the available amounts of speech in regional dialects should be made available for ASR, by providing phonetic transcripts instead of analogous translations in standard German. Finally, datasets for L2 speakers as well as those containing technical terms, e.g., in the medical or legal domain would be desirable.

By applying a high-quality (neural) language model, we expect a further decrease in WER of 1-2% absolute. We will investigate the training of an own language model since we did not find any high-quality model to be freely available.

Another aspect for future research is to evaluate the models' inferencing speed on different hardware to meet the demand of Aksënova et al. for paying attention to real-time factor and resource usage.

.